\documentclass[10pt,twocolumn,letterpaper]{article}

\usepackage{cvpr}              

\usepackage{graphicx}
\usepackage{xcolor}
\usepackage{svg}
\usepackage{bbding}
\usepackage{pifont}

\PassOptionsToPackage{dvipsnames}{xcolor}
\usepackage{xcolor}

\definecolor{cvprblue}{rgb}{0.21,0.49,0.74}
\usepackage[pagebackref,breaklinks,colorlinks,citecolor=cvprblue]{hyperref}
\newcommand{\xmark}{\text{\ding{55}}}
\def\paperID{*****} 
\def\confName{CVPR}
\def\confYear{2024}

\title{ControlEchoSynth: Boosting Ejection Fraction Estimation Models via Controlled Video Diffusion}

\author{
Nima Kondori\textsuperscript{1} \quad
Hanwen Liang\textsuperscript{2} \quad
Hooman Vaseli\textsuperscript{1} \quad
Bingyu Xie\textsuperscript{1} \quad
Christina Luong\textsuperscript{3} \\
Purang Abolmaesumi\textsuperscript{1} \quad
Teresa Tsang\textsuperscript{3} \quad
Renjie Liao\textsuperscript{1} \\[1em]
\textsuperscript{1}The University of British Columbia\\
\textsuperscript{2}University of Toronto\\
\textsuperscript{3}Vancouver General Hospital, Vancouver, BC, Canada\\
{\tt\small \{nimakondori, hoomanv, purang, rjliao\}@ece.ubc.ca} \\
{\tt\small hw.liang@mail.utoronto.ca} \\
{\tt\small \{christina.luong, t.tsang\}@ubc.ca}
}

\begin{document}
\maketitle
\begin{abstract}

Synthetic data generation represents a significant advancement in boosting the performance of machine learning (ML) models, particularly in fields where data acquisition is challenging, such as echocardiography. The acquisition and labeling of echocardiograms (echo) for heart assessment, crucial in point-of-care ultrasound (POCUS) settings, often encounter limitations due to the restricted number of echo views available, typically captured by operators with varying levels of experience. This study proposes a novel approach for enhancing clinical diagnosis accuracy by synthetically generating echo views. These views are conditioned on existing, real views of the heart, focusing specifically on the estimation of ejection fraction (EF)—a critical parameter traditionally measured from biplane apical views. By integrating a conditional generative model, we demonstrate an improvement in EF estimation accuracy, providing a comparative analysis with traditional methods. Preliminary results indicate that our synthetic echoes, when used to augment existing datasets, not only enhance EF estimation but also show potential in advancing the development of more robust, accurate, and clinically relevant ML models. This approach is anticipated to catalyze further research in synthetic data applications, paving the way for innovative solutions in medical imaging diagnostics.
\end{abstract}    
\section{Introduction}
\label{sec:intro}

Ejection fraction (EF) represents the proportion of the blood ejected by the heart during each heartbeat relative to its filled volume. It is a critical measurement for heart failure~\cite{huang2017ef}. Clinically, EF is measured using biplane Simpson's method, which involves two standard echocardiographic (echo) views of the heart: Apical 4-Chamber (A4C) and Apical 2-Chamber (A2C). The process involves measuring left ventricle (LV) volume in the End-Diastole (ED) and End-Systole (ES) phases and measuring the percentage of the volume pumped by the heart during the heartbeat. Echo is widely recognized as the preferred modality for measuring EF.
Nevertheless, a certain level of expertise is required to capture the data effectively. An expert clinician can use echo to identify the ED and ES frames and estimate the EF. Although it is possible to estimate EF from either A2C or A4C views, the biplane method, combining the measurements from both views, produces a more reliable and robust estimation of the EF~\cite{jafari2019,bras2019simpsons}. 
However, for the biplane method, obtaining samples in A2C view is comparatively more challenging for novice operator compared with A4C view.

Currently, the importance of synthetic data in the realm of machine learning for healthcare cannot be overstated. It serves as a vital tool to overcome the significant hurdle of data scarcity that has been impeding progress in this field, especially when compared to areas like computer vision, where data are more abundant. Synthetic data, which closely replicates the complexity and high-dimensionality of real-world medical data, allow researchers to potentially bypass privacy restrictions and access a wealth of information for their studies~\cite{Goncalves2020}. This not only accelerates the pace of innovation but also ensures that these advancements translate into real-world improvements in patient care. Therefore, the use of synthetic data is not just beneficial but essential for the growth and evolution of machine learning in healthcare. Without it, the progress in this field would likely continue to lag behind other sectors. Hence, synthetic data are potentially key catalyst in bridging this gap and propelling healthcare into a new era of data-driven, personalized medicine.

This study specifically tackles the scarcity of A2C echo sequences, proposing a novel methodology that employs a controlled video diffusion model to generate high-fidelity A2C echoes from A4C inputs. Our comprehensive evaluation reveals that this approach not only meets but exceeds current benchmarks in generating authentic data, and enhancing EF prediction models through data augmentation. An important advantage of our approach over prior efforts lies in our model's ability to operate without the necessity for ground truth segmentation of data. This aspect is particularly crucial in the medical domain, where the creation of these segmentation maps demands substantial time and resources. Finally, our findings indicate a notable improvement in EF model accuracy, underscoring the potential of controlled video diffusion models to mitigate data limitations in the medical field and contribute to the advancement of cardiac healthcare technologies.

\begin{figure*}
\centering 
\includegraphics[width=0.8\linewidth]{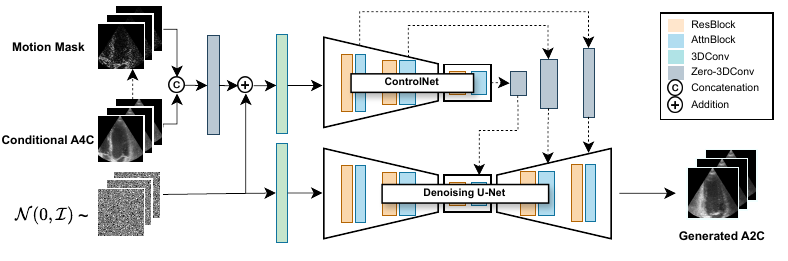}
\caption{\textbf{Model Overview}: Initially the denoising U-Net~\cite{unet} is trained unconditionally to generate A2C videos. The ControlNet branch is then added with the 3D zero convolution layers, allowing the model to generate conditional videos given the A4C echoes. The A4C conditioning is a channel-wise concatenation of the A4C echo with the motion mask generated from the A4C echo.}
\label{model_overview}
\end{figure*}

\section{Related Work}
\label{rel_work}
\subsection{Ejection Fraction}
With the rise of deep learning (DL), many vision-based works have tried to estimate the EF automatically. Zhang et al.~\cite{zhang2018fully} reported a fully automated echo interpretation system with convolutional neural networks (CNNs). Numerous studies within this domain are engaged in the segmentation of cardiac chambers. For instance, Liu et al.~\cite{liu2021deep} employ feature pyramids and a segmentation coherency network to perform LV segmentation. Cheng et al.~\cite{cheng2022contrastive} and Thomas et al.~\cite{thomas2022light}, on the other hand, leverage contrastive learning and Graph Neural Networks (GNNs) to achieve similar objectives in cardiac chamber segmentation. Recently, Mokhtari et al.~\cite{mokhtari2022echognn} provide explainability in their framework by learning a graph structure among frames of an echo.

\subsection{Diffusion Models}
Diffusion models are a class of generative models that learn to generate data by modeling the process of gradually adding noise to data and then learning to reverse this process. Recently, Latent Diffusion Models (LDMs) ~\cite{rombach2022high} perform the time-consuming diffusion process in latent space and it is further extended to Stable Diffusion~\cite{stablediff} by training on large-scale text-image datasets.  However, a notable limitation inherent in these models is their diminished capacity for controlled generation. To enhance the controllability of the diffusion process, unCLIP~\cite{ramesh2022hierarchical} extracts CLIP features from the text and then concatenates the CLIP feature during the diffusion steps or uses a cross-attention module. SDEdit~\cite{meng2021sdedit} achieves controllable image editing by adding noise to the given stroke without extra training steps for diffusion models. 
ControlNet~\cite{zhang2023adding} is proposed for adding extra conditions (e.g., canny map, depth map, segmentation map, etc.) to pre-trained diffusion models, which makes it able to control the content structure of the generated results without sacrificing the generation ability. 
Inspired by the success of text-to-image generation, several works are proposed for generating videos based on diffusion models. Video diffusion models~\cite{ho2022video} used 3D U-Nets to extend DMs to the domain of video generation. Several other notable works, including Imagen Video~\cite{ho2022imagen}, Make-A-Video~\cite{singer2022make}, and Animatediff~\cite{guo2023animatediff}, extend text-to-image diffusion models by training them on extensive text-video pairs. Furthermore, SVD~\cite{blattmann2023stable} has achieved impressive results by expanding LDMs to work with video data.

A series of interesting works in the medical field have shown promising results in echocardiography field. For instance,
Gu et al.~\cite{gu2021echocardiogram} use Generative Adversarial Networks (GANs)~\cite{goodfellow2014generative} to generate the A2C view conditioned on the A4C view using the segmentation map as conditioning on their generation. Nguyen et al.~\cite{van2023echocardiography} employ diffusion models, conditioned on the segmentation labels of the ED frame, to synthesize realistic echo videos corresponding to the segmentation map. Pellicer et al.~\cite{Pellicer2023.11.11.566718} employ diffusion models and a super-resolution GAN to synthesize unconditional high-resolution echoes. Reynaud et al.~\cite{Reynaud_2023} explore the reconstruction and counterfactual generation of echoes using cascaded diffusion models~\cite{ho2021cascaded} conditioned on an initial frame and an EF value. The generated data is used to boost the performance of an EF estimation model; however, their model is limited to same view generation.
\section{Methodology}
\label{sec:methodology}
In this section, we provide a detailed illustration of the proposed generation method. We begin by introducing preliminary works on diffusion models. Subsequently, we delve into the components of our proposed echo synthesis model to give a more detailed introduction. Later, we discuss the application of our proposed diffusion model to enhance the downstream EF prediction task, outlining its implementation and benefits.
\subsection{Preliminary}
\paragraph{\textbf{Vanilla Diffusion Models}}
For a general unconditional diffusion model, given an input signal $x_{0}$, a diffusion forward process is defined as
\begin{equation}
q_{\theta}(x_{t}\mid x_{t-1})= \mathcal{N}(x_{t};\sqrt{1-\beta_{t-1}}x_{t-1},\beta_{t}\mathit{I}),   t=1,...,T 
\label{forward_process}
\end{equation}
where $T$ is the total time step of the diffusion process. A noise depending on variance $\beta_{t}$ is gradually added to $x_{t-1}$ to obtain $x_{t}$ at the next time step and finally reach $x_{T}\in \mathcal{N}(0,I)$. The goal of the diffusion model is to learn to reverse the diffusion process, which is called the denoising process. Given a random noise $x_{t}$, the model predicts the added noise at the next time step $x_{t-1}$ until reaching the origin signal $x_{0}$:
\begin{equation}
p_{\theta}(x_{t-1}\mid x_{t})= \mathcal{N}(x_{t-1};\mu_{\theta}(x_{t},t),{\textstyle \sum}_{\theta}(x_{t},t)),   t=T,...1 .
\end{equation}
We fix the variance ${\textstyle \sum}_{\theta}(x_{t},t)$ and utilize the diffusion model with parameter $\theta$ to predict the mean of the inverse process $\mu_{\theta}(x_{t},t)$. The model can be simplified as denosing models $\epsilon_{\theta}(x_{t},t)$, which are trained to predict the noise of $x_{t}$ with a noise prediction loss:
\begin{equation}
min_{\theta}{\left\|\epsilon-\epsilon_{\theta}(x_{t},t)\right\|}_{2}^{2} ,
\label{diff_loss}
\end{equation}
where $\epsilon$ is the noise added to the source image $x_{0}$, and this loss function is for unconditional diffusion model. When some extra conditions, e.g. the text prompt $c_{p}$, are added, the above loss function is reformulated as 
\begin{equation}
min_{\theta}{\left\|\epsilon-\epsilon_{\theta}(x_{t},t,c_{p})\right\|}_{2}^{2} ,
\label{cond_diff_loss}
\end{equation}
where the model learns to predict the noise of $x_{t}$ conditioned on text prompt $c_{p}$ at time step $t$.  

\paragraph{\textbf{ControlNet}}
ControlNet~\cite{zhang2023adding} is a neural network architecture that enhances pre-trained diffusion models with task-specific conditions by utilizing trainable layers copied from the original diffusion model. 
It adds several zero-convolution layers on the top, and these layers are then fine-tuned together based on specific control signals such as edge, depth, and segmentation inputs~\cite{zhang2023adding}. The loss with additional control can be formulated as
\begin{equation}
min_{\theta}{\left\|\epsilon-\epsilon_{\theta}(x_{t},t,c_{p},c_{f})\right\|}_{2}^{2} ,
\end{equation}
where $c_f$ is the additional control. Our method draws inspiration from ControlNet and expands its application into video synthesis.

\subsection{Controlled Echo Synthesis}
In this study, we target at generating the A2C views conditioned on the A4C views. In the case of a vide, the input signal $x$ and control $c_f$ are extended from single images to video clips. 
Specifically, given a pair of echo videos $x_{A4C}$ and $x_{A2C} \in \mathbb{R}^{T \times C \times H \times W}$, where $T$, $C$, $H$, and $W$ denotes the number of frames, number of channels, height, and width, we take $x_{A4C}$ view as a condition and develop a controlled echo synthesis network to diffuse the $x_{A2C}$ from random noise $x_{T}$. 

The overall framework, as shown in \cref{model_overview}, is composed of a main U-Net branch and a ControlNet branch which is trained to generate the ground truth A2C echo sequence corresponding to the condition. 
As most of the open-source video-based latent diffusion models are pre-trained on large-scale natural videos rather than echocardiographical images, we circumvent the domain difference problem and train a U-Net model from scratch in the RGB space.
Initially, the main U-Net branch undergoes training from scratch in an unconditional manner on AC2 views. In the forward process, $x_{A2C}$ is noised following Eq.~\ref{forward_process} with T=1000 steps and the model learns to reverse the forward steps through the denoising U-Net~\cite{unet}.
After train the U-Net unconditionally in the first phase, we adopt it to the second phase of conditional training. 
Inspired by~\cite{zhang2023adding}, in this phase, we add the ControlNet branch to digest the conditions. 
The ControlNet branch is generated by copying the encoder and the middle blocks of the denoising U-Net weights to the control branch. We use another three Zero-3DConv layers to construct the decoder part of ControlNet, whose outputs are further fed into the Denoising U-Net.
To enhance the model's sensitivity to motion—particularly crucial for the targeted downstream task—we refine the motion priors associated with the conditional A4C inputs. This is achieved by calculating the motion across A4C frames through the subtraction of pixel intensities between consecutive frames, followed by the application of Gaussian smoothing. This technique accentuates regions within the echo sequence exhibiting significant motion, thereby providing a focused analysis for more effective EF prediction.
The motion mask, generated through this process, is concatenated along the channel dimension with the original A4C video clip. Subsequently, a Zero-3DConv layer is added atop the encoder section, facilitating dynamic adjustments throughout the training phase, ensuring the model incrementally adapts to the intricacies of motion within the echocardiographic data.

\subsection{EF Prediction}
\label{method:ef}
In this study, we assess the efficacy of our diffusion model and the utility of synthetic echo videos in the downstream task of EF estimation. 
We formulate this as a regression problem, directly predicting EF from echo videos without requiring LV chamber segmentation.
To comprehensively evaluate the generalizability of our synthetic data across different architectural paradigms, we employ two distinct models: one based on transformer architecture and another on convolutional neural network (CNN) architecture.
Specifically, we employ ResNet2+1D~\cite{r2p1d}, a widely recognized CNN-based model in video analysis, and previously utilized by Reynaud et al. as their EF model~\cite{Reynaud_2023} in assessing their proposed diffusion model.
Additionally, we incorporate EchoCoTr-S~\cite{EchoCoTr}, the state-of-the-art transformer-based architecture for EF estimation, which is built on UniFormer~\cite{li2022uniformer}.
We configure our EF models in two modes: single-plane, where EF is predicted solely from the patient's A4C echo view, and biplane, where EF is inferred by averaging the predictions from both A2C and A4C echo views.
Throughout both configurations, we optimize the EF models using the mean squared error (MSE) as the loss function.

\section{Experiments}

\begin{figure*}[!h]
\centering 
\includegraphics[width=\linewidth]{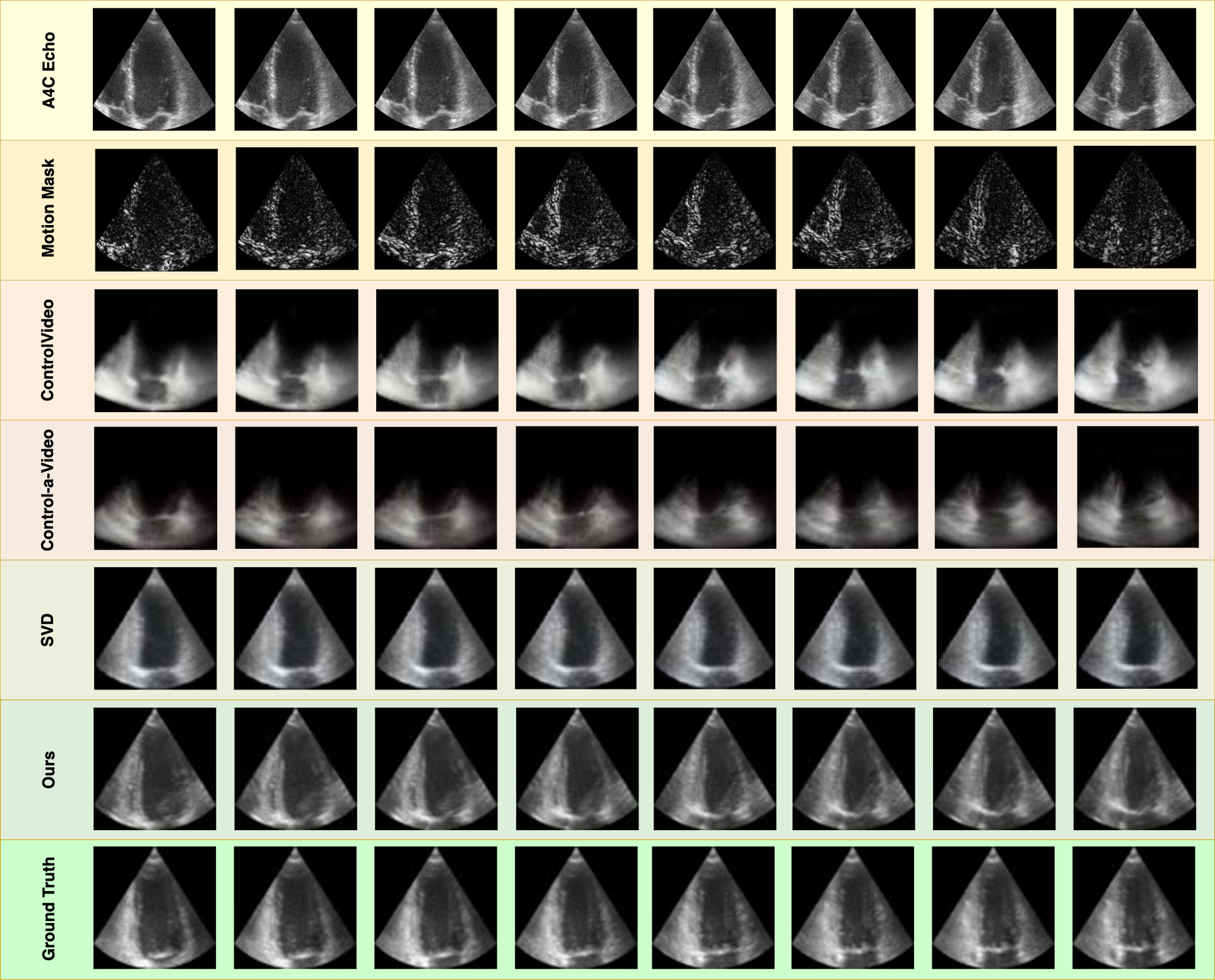}
\caption{\textbf{Qualitative Analysis}: Comparison of the quality of the generation of different models. The top row shows the A4C echo concatenated in the channel dimension with the motion mask frames from the second row. The last row shows the A2C frames of the patient. As can be seen from the prediction frames, our model captures the shape and the motion of LV from the provided conditioning more accurately compared to the previous work, generating a realistic and consistent A2C video.} \label{qualitative_analysis}
\end{figure*}

\begin{table*}
  \centering
  \begin{tabular}{llcccccc}
    \toprule
    Model Backbone & Training Dataset & R2$\uparrow$ & MAE$\downarrow$  & RMSE$\downarrow$  \\
    \midrule
    \midrule
    ResNet2+1D~\cite{r2p1d}            & A4C                        & 0.623          & 4.58          & 6.25           \\
    ResNet2+1D~\cite{r2p1d}            & A4C \& A2C                 & 0.677          & 4.43          & 5.78           \\
    ResNet2+1D~\cite{r2p1d}        & A4C \& Synth A2C           & \textbf{0.713} & \textbf{3.85} & \textbf{5.46}  \\
    ResNet2+1D~\cite{r2p1d}        & A4C \& A2C \& Synth A2C    & 0.662	       & 4.60	       & 5.92           \\
    \hline
    EchoCoTr-S~\cite{EchoCoTr}         & A4C                        & 0.231          & 6.90           & 8.93           \\
    EchoCoTr-S~\cite{EchoCoTr}         & A4C \& A2C                 & 0.554          & 5.28          & 6.80           \\
    EchoCoTr-S~\cite{EchoCoTr}       & A4C \& Synth A2C           & \textbf{0.642} & \textbf{4.75} & \textbf{6.09} \\
    EchoCoTr-S~\cite{EchoCoTr}       & A4C \& A2C \& Synth A2C    & 0.614          & 4.83          & 6.32           \\
    \bottomrule
  \end{tabular}
  \caption{We compare the performance of the model trained on different training datasets, including real and synthetic data. All the metrics are measured against the real A2C and A4C echo videos in the test set.}
  \label{tab:EF}
\end{table*}

\subsection{Dataset}
\label{sec:dataset}
For our dataset, we take advantage of the CAMUS dataset~\cite{leclerc2019deep} that includes a train set of 450 pairs and a test set of 50 pairs of A4C and A2C views. We train our controlled echo synthesis model on the training set of CAMUS and evaluate it on the test set. Furthermore, in our experiments, we also take advantage of an internal biplane dataset consisting of 8,074 pairs of A4C and A2C echos. The A2C echos are used in the first unconditional training phase to train our denoising U-Net. 

\begin{table*}[!htpb]
  \centering
  \begin{tabular}{lccllll}
    \toprule
    Model & Pre-trained & Backbone Finetuned & FVD$\downarrow$ & FID$\downarrow$ & SSIM$\uparrow$ & LPIPS$\downarrow$ \\
    \midrule
    \midrule
    Using CAMUS only & \xmark      & \xmark     & 72.38          & 20.40          & 0.50          &  0.18    \\
    From scratch     & \xmark      & \checkmark & 99.80          & \textbf{18.97} &  0.49         &  0.20    \\ 
    Frozen U-Net     & \checkmark  & \xmark     & 72.14          & 29.55          & 0.56          &  0.17    \\
    Ours             & \checkmark  & \checkmark & \textbf{69.58} & 26.64          & \textbf{0.57} &  \textbf{0.16} \\ 
    \bottomrule
  \end{tabular}
  \vspace{-0.1cm}
  \caption{\textbf{Ablations}: We compare different training paradigms of our model. The first row refers to training the entire model from scratch. The second row was the multi-step training of unconditional and control branches on CAMUS dataset. The third row refers to freezing the U-Net in the second stage of training while using the U-Net model trained on the internal dataset. The last row refers to our model where U-Net is trained on the internal data and the entire model is trained further on the CAMUS dataset.}
  \label{tab:ablation}
  \vspace{-0.3cm}
\end{table*}

\begin{table}[!htb]
\vspace{0.15cm}
  \centering
  \begin{tabular}{lllll}
    \toprule
    Model & FVD$\downarrow$ & FID$\downarrow$ & SSIM$\uparrow$ & LPIPS$\downarrow$\\
    \midrule
    \midrule
    ControlVideo~\cite{zhang2023controlvideo}    & 200.50 &	33.25 & 0.24 &     0.47    \\
    Control-a-Video~\cite{chen2023controlavideo}    & 181.90 &	31.99 & 0.26  &    0.41      \\
    SVD~\cite{blattmann2023stable}        & 154.94          & 31.43 &  0.32 &  0.35                         \\
    SPADE~\cite{van2023echocardiography}    & 89.78           & \textbf{18.65} & 0.56 &                               \\
    Ours   & \textbf{69.58}  & 26.64 & \textbf{0.57} & \textbf{0.16} \\ 
    \bottomrule
  \end{tabular}
  \caption{\textbf{Quantitative Analysis}: Comparison of the quality of the generated videos through the relevant metrics.}
  \vspace{-0.4cm}
  \label{tab:diff_fvd}
\end{table}

\subsection{Implementation Detail}
\label{sec:implementation}
\subsubsection{Diffusion Model}
 
 For the U-Net training, we resized A2C echo clips to $64 \times 64$ spatially. We use a sliding window of the size 32 frames across the video and use downsampling of the factor 2 on the sliding window to create 16 frames. The input to our forward diffusion process $x_{A2C} \in \mathbb{R}^{16\times 64 \times 64}$ is then gradually noised as explained in \cref{sec:methodology} and the U-Net is trained to deniose it. We used a cosine annealing learning rate scheduler with a maximum of $1e-4$ and a minimum of $1e-7$. Each level in the denoising U-Net consists of ResNet \cite{resnet} blocks followed by temporal attention \cite{vaswani2017attention} layers and involves scaling (upsampling or downsampling) of dimensions. After completing the initial training phase, we paused the process to integrate control branches into the model. Following~\cite{zhang2023adding}, we created a copy of the encoder and the middle part of our denoising U-Net and, using a Zero-3DConv layer, added the output of our control branch to each layer in the U-Net branch. Unlike typical ControlNet training approaches, we empirically discovered that freezing the U-Net branch resulted in a slight degradation of video quality; hence, we opted to fine-tune the U-Net branch alongside the control branch. The model then underwent an additional 80,000 iterations with the U-Net branch being trainable, employing a cosine annealing learning rate of $5e-5$ with a warmup of 10 iterations. Comparative results across various training adaptations of our model are presented in \cref{tab:ablation}.

\subsubsection{EF Model}
For training and evaluating the EF models, we exclusively utilized the CAMUS dataset.
From the total of 450 training studies, we randomly allocated 50 studies as a validation set for hyperparameter tuning.
The selection of optimal models and hyperparameters was contingent upon the Mean Squared Error (MSE) value computed on this validation set.
Echo frames were resized to dimensions of $64\times64$ pixels using bilinear interpolation.
To standardize the temporal aspect, we resized the video to consist of 16 frames by selecting the nearest neighbor frames from a sliding window of 32 frames.
Hyperparameter tuning was facilitated using the sweep functionality provided by the online platform Weights and Biases~\cite{wandb}.
Specifically, we explored the hyperparameter space encompassing batch size ($8$, $16$, $24$), number of training epochs ($50$, $100$, $150$), learning rate ($1e-3$, $5e-4$, $1e-4$), and learning rate scheduler (Reduce on Plateau, Step, Cosine Annealing). 
Our experiments and developments were conducted utilizing the PyTorch library~\cite{Pytorch} and one V100 16GB GPU.

\subsection{Quantitative Results}
In terms of the generative capabilities of our model, in ~\cref{tab:diff_fvd}, we show that our model achieves the best results in terms of FVD, SSIM, and LPIPS metrics and achieves a decent FID score.
The results of our EF models on the CAMUS dataset's test set are summarized in ~\cref{tab:EF}.
As detailed in ~\cref{method:ef}, in the biplane mode, the EF models receive input as A4C-A2C pairs of echo videos.
We present biplane results across three configurations:
utilizing solely real data (comprising only real A4C-A2C views),
synthetic data (incorporating real A4C with synthetic A2C views),
and an augmented dataset (combining both real and synthetic A2C views).
Synthetic A2C echo videos were selected through a rigorous curation process. 
After generating 18 synthetic A2C echos per patient, conditioned on A4C echos, we estimated their EF using a pre-trained EF model.
Subsequently, the synthetic A2C echo videos were ranked based on their absolute error from the ground truth EF of the conditioning A4C video.
To ensure the inclusion of relevant and EF-realistic synthetic A2C videos, we selected the top three echos from the ranking.
These selected synthetic A2C videos were then paired with their respective conditioning A4C videos, contributing to the synthetic dataset.
Consequently, this dataset offered three times the training cases of the real dataset, augmenting the training process for improved EF model performance.
Our evaluation metrics included R2 score, mean absolute error (MAE), and root mean squared error (RMSE). 
The results indicate that biplane models outperform single-plane models.
Furthermore, both model architectures exhibited superior performance across the board when trained solely on synthetic A2C echos.  
We hypothesize that the training the U-Net on our internal larger dataset may have enhanced the quality of the synthetic cases, thereby improving EF model performance, especially when compared to training on the augmented dataset, in which real cases may be more difficult to train on.
Moreover, the EchoCoTr-S, a transformer-based model, benefits notably from the augmented dataset, potentially due to its reliance on a larger training dataset, in contrast to ResNet2+1D, which demonstrates high performance even with the real dataset alone.

\subsection{Qualitative Results}
As can be seen in~\cref{qualitative_analysis}, our model performs significantly better than the previous works not only in terms of the quality of the generated data but also in terms of capturing the shape and motion of the LV which is essential in accurately estimating the EF of the heart.

\subsection{Ablation Studies}
We show the results of the ablation studies performed on our diffusion model. As can be seen from ~\cref{tab:ablation}, training the U-Net model on the internal dataset plays an important role in the improvement of the quality of the generated data. We also can see how multi-step training of the model results in better quality metrics than training the entire model from scratch. We also observe that using the model trained on the internal dataset introduces a large improvement in FID, SSIM, and LPIPS scores. It is also important to observe that freezing the U-Net model does not perform as well as tuning it in the training of our conditional model.

\section{Conclusions}
In this paper, we are exploring the advantages of generating synthetic data to boost the performance of ML models in clinical settings. Our contribution lies in the proposition of a novel controlled conditional video generation model, designed to facilitate the augmentation of available data resources, particularly in scenarios where data scarcity poses a significant concern. The conditioning of the model can be tailored to the downstream task. This adaptability ensures that the synthetic data produced exhibits a heightened relevance and utility in addressing the intricacies of the intended applications within clinical settings. Additionally, we show the model agnostic power of our approach by measuring the EF estimation performance boost due to the synthetic data on two different model architectures, namely ResNet2+1D~\cite{r2p1d} and EchoCoTR~\cite{EchoCoTr}. 
\section{Future work}
While our project does not aim to generate an accurate A2C echo solely based on the A4C view, we recognize the potential for enhancing control by incorporating additional patient metadata or other views. This could lead to more precise outputs. Moreover, the scalability of our model opens up avenues for future exploration, such as integrating controls for depth, brightness, and other parameters. These advancements could further refine the synthetic data generation process and expand its applicability in diverse contexts.
{
    \small
    \bibliographystyle{ieeenat_fullname}
    \bibliography{main}
}


\end{document}